\documentclass[final, 3p, times]{elsarticle}
\usepackage{epstopdf}
\usepackage{caption}
\usepackage{setspace}
\usepackage{amssymb}
\usepackage{multirow}
\usepackage{amsthm}
\usepackage{booktabs}
\usepackage{amsmath}
\usepackage{amssymb}
\usepackage{mathrsfs}
\usepackage{longtable}
\usepackage{lscape}
\usepackage{url}
\usepackage{tabularx}
\usepackage{epsfig}
\usepackage{colortbl}
\usepackage{subfigure}
\usepackage{tikz,mathpazo}
\usepackage{graphicx}
\usepackage{natbib}
\usepackage[noend]{algpseudocode}
\usepackage{algorithm}
\usepackage{algorithmicx}   
\usepackage{array}  
\usepackage{hyperref}
\usepackage[figuresright]{rotating}

\newtheorem{myDef}{Definition}

\usetikzlibrary{shapes.geometric, arrows}
\begin{document}
	\begin{frontmatter}
		\title{DVS: Deep Visibility Series and Application in Construction Cost Index Forecasting}
		\author[address1,address2]{Tianxiang Zhan \corref{myauthor}}
		\author[address1,address2]{Yuanpeng He \corref{myauthor}}
		\author[address1,address2]{Hanwen Li \corref{myauthor}}
		\author[address2,address3]{Fuyuan Xiao \corref{mycorrespondingauthor}}
		
		\address[address1]{College of Computer and Information Science College of Software, Southwest University, Chongqing, 400715, China}
		\address[address2]{School of Big Data and Software Engineering, Chongqing University, Chongqing, 401331, China}
		\address[address3]{National Engineering Laboratory for Integrated Aero-Space-Ground-Ocean Big Data Application Technology, China}
		\cortext[myauthor]{The e-mail of Tianxiang Zhan: zhantianxiangswu@163.com; The e-mail of Yuanpeng He: heyuanpengswu@163.com; The e-mail of Hanwen Li: lihanwenswu@163.com
		}
		\cortext[mycorrespondingauthor]{Corresponding author: Fuyuan Xiao is with the School of Big Data and Software Engineering, Chongqing University, Chongqing 401331, China. (e-mail: xiaofuyuan@cqu.edu.cn; doctorxiaofy@hotmail.com)
		}

		\begin{abstract}
	Time series forecasting is a hot spot in recent years. Visibility Graph (VG) algorithm is used for time series forecasting in previous research, but the forecasting effect is not as good as deep learning prediction methods such as methods based on Artificial Neural Network (ANN), Convolutional Neural Network (CNN) and Long Short-Term Memory Network (LSTM). The visibility graph generated from specific time series contains abundant network information, but the previous forecasting method did not effectively use the network information to forecast, resulting in relatively large prediction errors. To optimize the forecasting method based on VG, this article proposes the Deep Visibility Series (DVS) module through the bionic design of VG and the expansion of the past research. By applying the bionic design of biological vision to VG, DVS has obtained superior forecasting accuracy. At the same time, this paper applies the DVS forecasting method to the construction cost index forecast, which has practical significance.
		\end{abstract}
	
	
	\end{frontmatter}
	
	\section{Introduction}
	Time series forecasting is a research hot spot in the fields of mathematics, economy and so on. Time series forecasting technology has made many contributions to society such as disease prevention, risk forecasting, economic planning and so on. It is the goal of researchers to improve the efficiency and reduce the error of time series forecasting. 
	
	 Some classic statistical models such as Autoregressive Integrated Moving Average Model (ARIMA) \cite{tseng2002combining} can be combined with new technologies such as neural networks to reduce forecasting errors and improve efficiency. Time series forecasting based on Visibility Graph (VG) is a classic time series forecasting method which has small forecasting error  \cite{zhang2017novel, xu2018novel, mao2019time}. However, due to the high time complexity of the VG algorithm combined with the random walk algorithm, effective prediction information cannot be fed back in time. To optimize the method based on VG, this paper combines VG and Convolutional Neural Network (CNN). The article proposes Enhanced Visibility Graph (EVG) and Deep Visibility Series (DVS) algorithm, which makes the visibility graph contains more information of network and rough time series, which improves the efficiency and further reduces the forecasting error.
	
	In the experiment, DVS method is used to predict a actual problem Construction Cost Index (CCI). The CCI is an indicator that reflects the construction cost published by Engineering News Record (ENR) once a month \cite{shahandashti2013forecasting,hwang2011time}. CCI forecasts are greatly meaningful in the construction and financial fields, and the classic data set of CCI time series forecasts effectively reflects the performance of forecasting methods.
	
	The structure of this paper is as follows: The second part introduces the VG algorithm and previous work. The third part will introduce how to optimize the VG algorithm. The fourth part will forecast the CCI data set, and compare it with statistical forecasting methods, hybrid forecasting methods, and machine learning methods, and conduct ablation experiments to prove the role of DVS. The last part summarizes the article and discusses future work.
	
	\section{Previous research}
	
	This section will introduce previous research on VG. Firstly, the definition of time series $T$ is as follows:
	\begin{equation}
		T = \left\{T_1,T_2,T_3,...,T_n\right\}
	\end{equation}

	\begin{equation}
	T_i = (t_i,v_i)
	\end{equation}
	where $t_i$ is the time of time point $T_i$ and $v_i$ is the value of the time point $T_i$.
	
	\subsection{Visibility Graph and forecasting method}
	
	\subsubsection{Visibility Graph}
	
	VG is a algorithm of converting a time series into a network. By inputting the time series, it can output an adjacency matrix that represents the connection relationship between the time series inspection points \cite{lacasa2008time}. 
	
	\begin{myDef}
		The visual relationship is a visual relationship constructed between two nodes in the time series based on the numerical value. The visual relationship has the same bionic characteristics as human vision. If an object is blocked behind an obstacle, humans cannot see the object, and there is no visual relationship between humans and the object. Humans can see obstacles, so obstacles are related to possession. So the definition of the relationship is as follows:

		\begin{equation}
		visual(T_i,T_j) = \begin{cases}
			0, & \mathop{\arg\max}\limits_{\theta \in [i,j],\theta \in Z} tan(T_\theta, T_j) \neq i\\
			1, &  \mathop{\arg\max}\limits_{\theta \in [i,j],\theta \in Z} tan(T_\theta, T_j) = i\\
		\end{cases}
	\end{equation}

 The Formula.4 here means that if the slope between nodes $i$ and $j$ is the maximum value of the slope between node $i$ and node $j$, then nodes $i$ and $j$ are visible, and the boolean state variable is 1, otherwise 0.
	\end{myDef}
	
	\begin{figure}[htbp]
		\centerline{\includegraphics[scale=0.5]{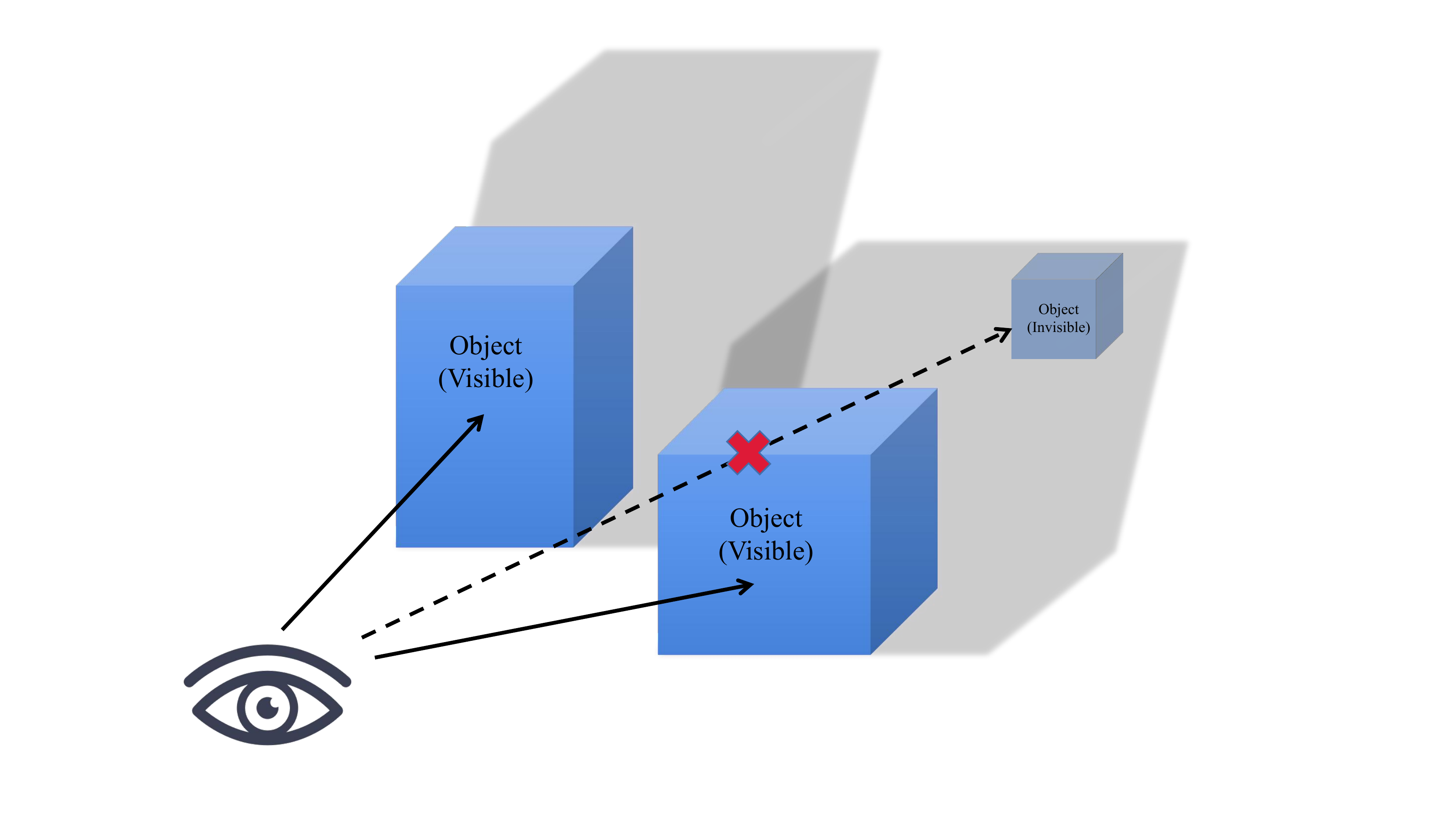}}
		\caption{Bionic example of the VG algorithm}
	\end{figure}
	
	Fig.1 is a bionic example of the VG algorithm. For the observer, observation can bargain two objects, but the blocked object cannot be seen. The observer only knows whether the object can be seen or not, and the properties of the object such as color cannot be seen. For time series, the property that the VG algorithm cannot represent is the value corresponding to the time node. The visual relationship can construct a rough adjacency matrix $A$ to represent the two-dimensional relationship in the entire time series.
	
	\begin{myDef}
		The definition of visual matrix $A$ is as follows:
		\begin{equation}
			A_{ij} = visual(T_i,T_j)
		\end{equation}
		where visual matrix $A$ contains the association relationship of the visual relationship.
	\end{myDef}
	
	Fig.2 below is an example of a VG. If the two nodes are visible, use the edge to connect the corresponding visual nodes. The connections between nodes in the time series are intuitively established.
	
	\begin{figure}[htbp]
		\centerline{\includegraphics[scale=0.9]{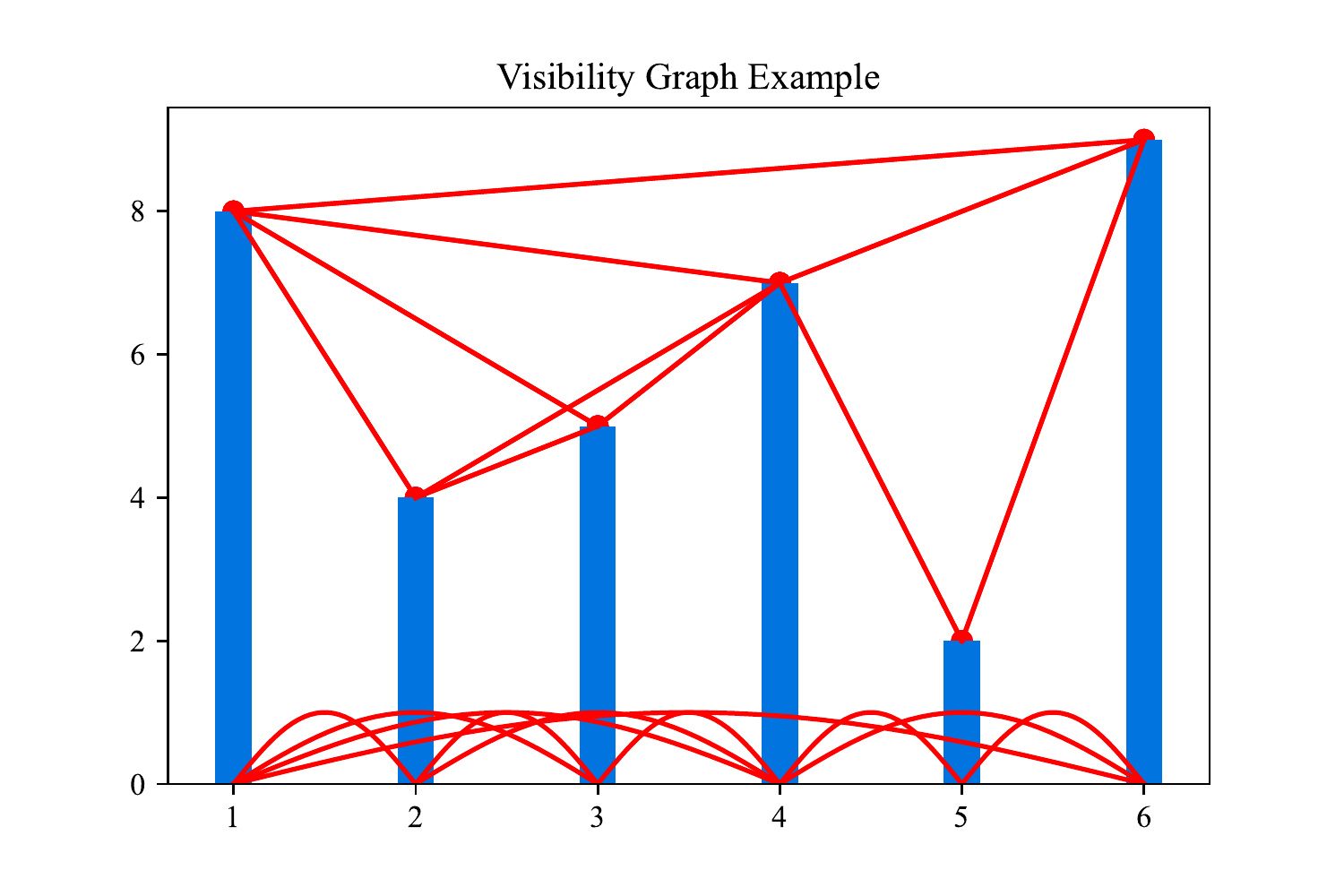}}
		\caption{Example of VG ($T = \left\{(1, 8),(2, 4),(3, 5),(4, 7),(5, 2),(6, 9)\right\}$)}
	\end{figure}
	
	\subsubsection{Forecasting based on Visibility Graph}
	
	In the previous research, after the time series was processed by the VG algorithm, the network generated by the VG was used to cooperate with the random walk algorithm. The purpose of the random walk algorithm is to find the similarity or the strength of the relationship between two nodes in the network \cite{liu2010link} . Finally, by selecting a node or a group of nodes with the highest similarity or relationship strength to fit the last time node linearly, the final result is obtained by weighting according to time distance, node similarity or other distances such as Fig.3.
	
	\begin{figure}[htbp]
		\centerline{\includegraphics[scale=0.9]{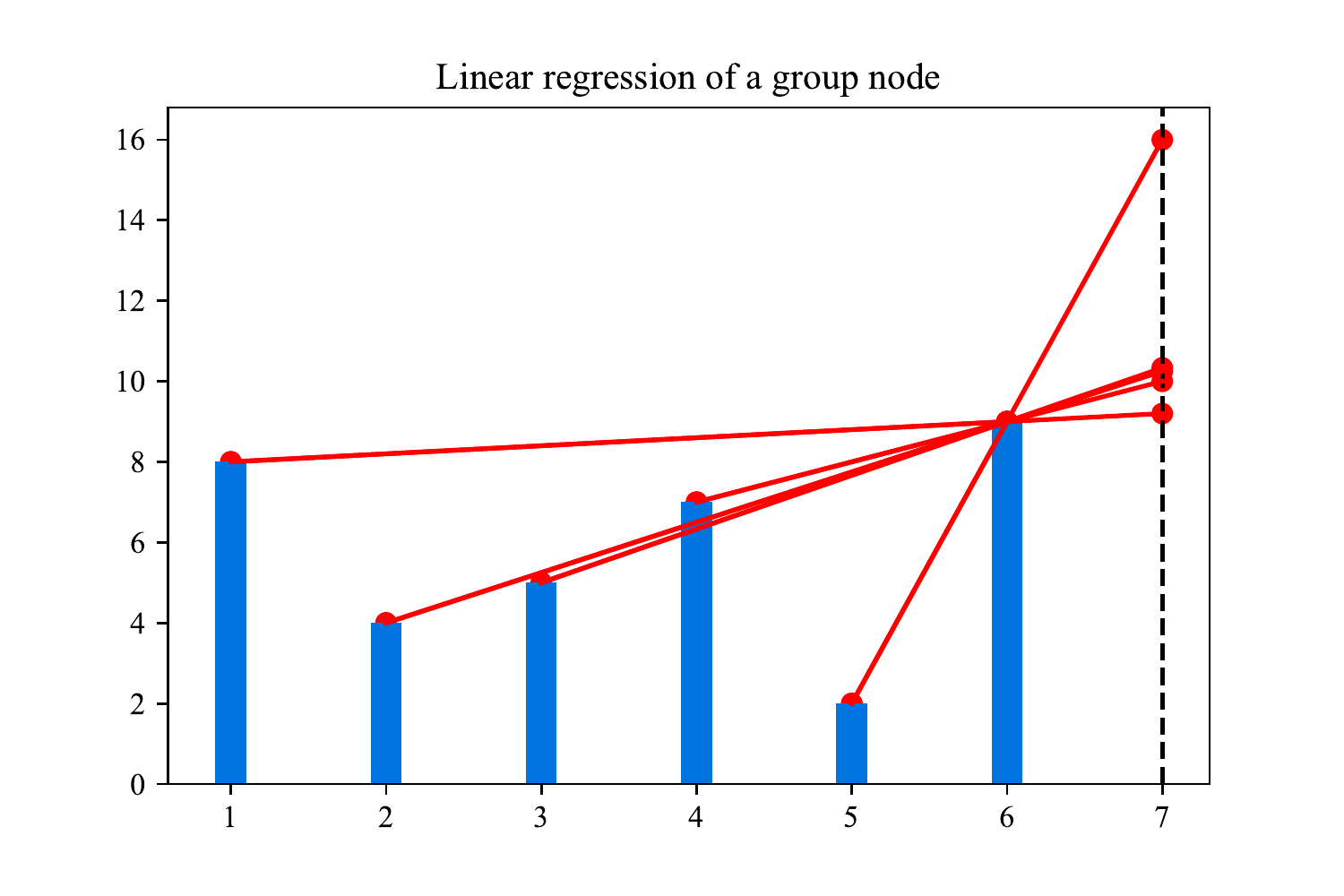}}
		\caption{Linear Regression of each node and last node}
	\end{figure}
	
	\subsection{How the previous research work}
	
	For the previous research, the forecasting algorithm is divided into the following three steps:
	\begin{enumerate}
		\item Generate a visibility graph of the time series.
		\item After random walk, find the point or a group of points with the highest similarity to the last node.
		\item After the selected set of points is linearly fitted to the last node, the weighted sum of the fitting results is used as the final forecasting result.
	\end{enumerate}
	
	Different from statistical linear fitting, the VG algorithm can be used to determine that some points are involved in the fitting instead of all points. At the same time, different time points are assigned a weight based on distance or similarity, which further reduces the forecasting error of the algorithm. The above analysis is the reason why the previous research is effective. For forecasting algorithms based on VG, how to efficiently use the nature of the network while retaining time series information is the core of the research.
	
	\subsection{Disadvantages of previous research}
	
	For the previous research, the feature of the network was preserved through the random walk algorithm, but in the end, only the time of the node was used, and other information was not used effectively. At the same time, the random walk algorithm requires a large number of cycles, and it stops when the state of the entire network tends to converge. In the end, only a few nodes in a row component of the adjacency matrix are used for forecasting. A large number of system resources are consumed in the process of random walk, and the information of the network is not used effectively.
	
	\section{Deep Visibility Series}
	
	This section will introduce the proposed DVS.
	
	\subsection{Enhanced visibility graph}
	
	For a time series $T$ and its adjacency matrix $A$ obtained by the VG algorithm, an enhanced visibility graph (EVG) algorithm is proposed. 
	
	\begin{myDef}
		EVG saves time series information in the network, and the output of EVG is an enhanced adjacency matrix $B$.
		
		\begin{equation}
			B_{ij} = \frac{A_{ij}*v_j}{Degree(T_i)}
		\end{equation}
		
		\begin{equation}
			Degree(T_i) = \sum_{j=1}^{n} A_{ij}
		\end{equation}
		where the meaning of the Degree function is the node degree of node $T_i$ in the VG network. Each element $B_{ij}$ in matrix $B$ represents the value at which node $T_i$ can see node $T_j$. Because node $T_i$ has chances to see more than one node, all the nodes it sees need to be divided by the degree of node $T_i$. Matrix $B$ also reflects the network characteristics of the time series while retaining the element information of the time series.
	\end{myDef}
	
	\begin{figure}[htbp]
		\centerline{\includegraphics[scale=0.5]{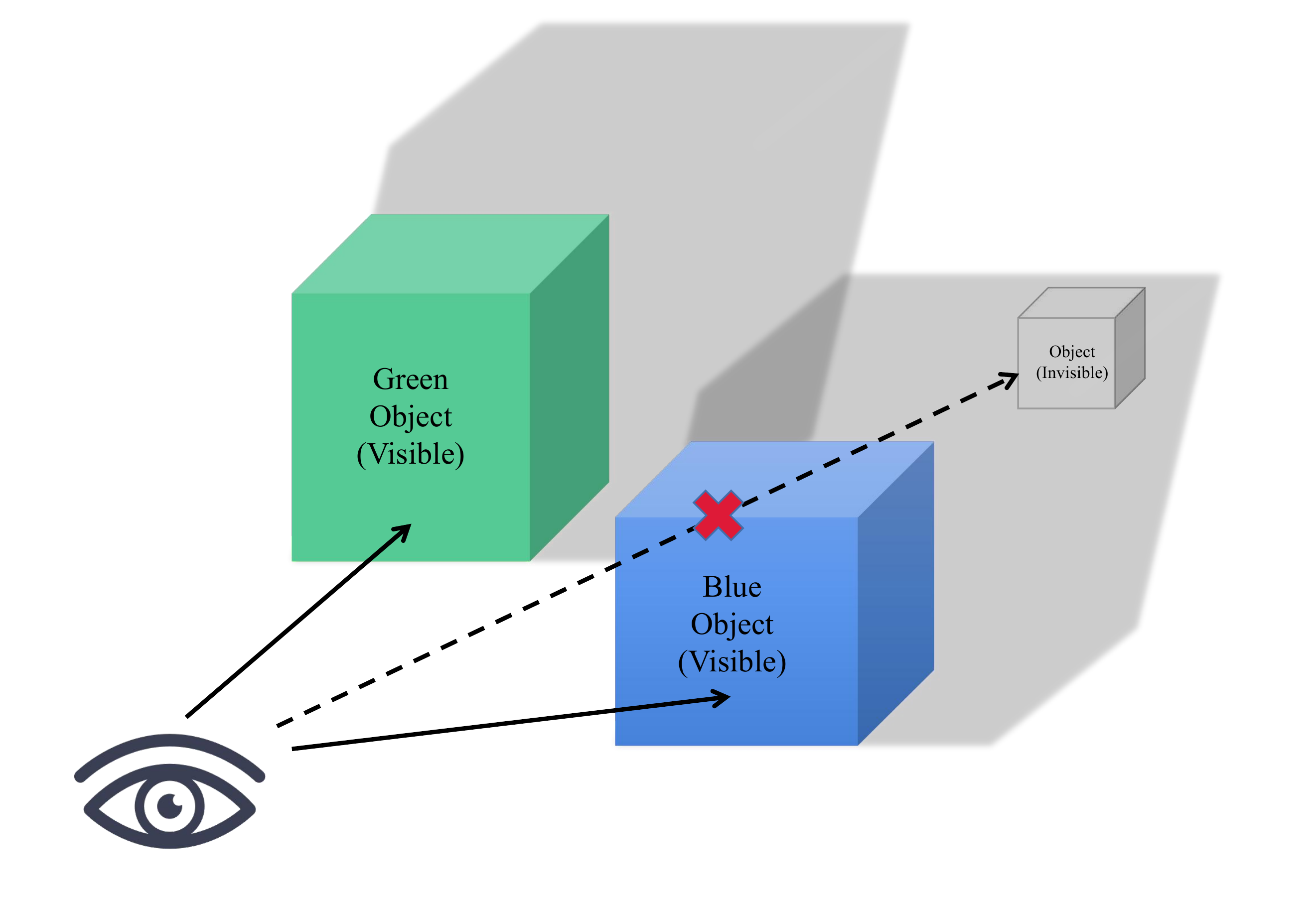}}
		\caption{Bionic example of the EVG algorithm}
	\end{figure}
	
	The EVG algorithm is also a bionic algorithm. Fig.4 is a bionic example of the EVG algorithm. For the observer, the observer can not only see the object but also the color of the object. The VG algorithm is equivalent to human beings knowing that I can see the object, but not knowing what the object is. EVG has increased the ability for humans to see exactly what an object is. Fig.5 is an example of the previous VG and time series conversion to EVG, and observer can recognize the color of objects.
	
	\begin{figure}[htbp]
		\centerline{\includegraphics[scale=0.5]{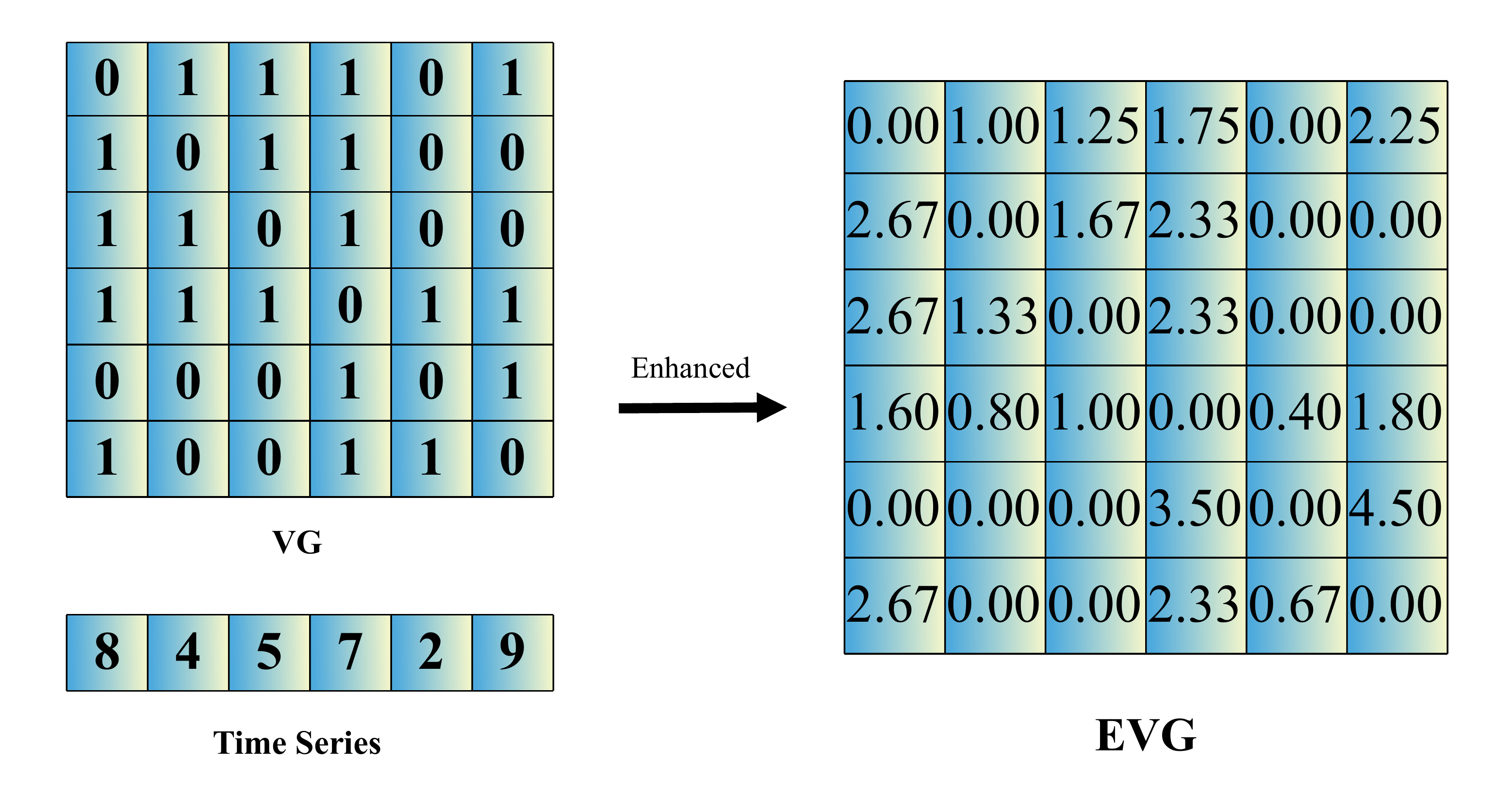}}
		\caption{Example of transform from time series and VG to EVG ($T = \left\{(1, 8),(2, 4),(3, 5),(4, 7),(5, 2),(6, 9)\right\}$)}
	\end{figure}
	
	\subsection{Deep Visibility Series Forecasting}
	Through the EVG algorithm, the problem of retaining network information is solved. Another problem is how to use the matrix generated by the existing EVG.
	
	\subsubsection{How to determine the form of information for forecasting}
	The matrix generated by EVG can be regarded as a picture. For pictures, a common operation is to use Convolutional Neural Network (CNN) or Artificial Neural Network (ANN) for picture recognition, which is a classification problem \cite{huang2021natural}. The time series forecasting problem is a non-discrete variable forecasting problem, and the matrix generated by EVG cannot be processed by means of pictures.
	
	Previous research has chosen to abandon most of the information in dealing with network information, allowing limited network information to be expressed in time series. The final forecasting operation is still the serialized information. On the other hand, the purpose of time series forecasting is not to over-strengthen the extraction of network information, but to enhance the expression of the sequence, so here is a way to compress the sequence generated by the EVG.
	
	\begin{myDef}
		When compressing the matrix generated by the EVG, the operation cannot be converted into a sequence such as Hash mapping by simply using a mapping method, because the original time series has three characteristics:
		\begin{enumerate}
			\item Orderliness: The order of the time series cannot be changed at will.
			\item Limited length: The number of time points is limited, one time corresponds to one value, and the number of values can only be equal to the number of times.
			\item Interpret ability: The value of each event has its own physical meaning and cannot be missing.
		\end{enumerate}
		Therefore, the compression algorithm uses the bionic idea here, which is defined as follows:
		\begin{equation}
			zip_i = \sum_{i=1}^{n} B_{ij}
		\end{equation}
		where matrix $B$ is the matrix generated by EVG. The $i-th$ element in the compressed sequence represents the sum of the visible values of the time node $T_i$. For humans, EVG distinguishes what each object is. 
	\end{myDef}
	
	Therefore, under the action of the EVG module and the ZIP module, the time series is converted into a time series that retains network information. Fig.6 shows the compression process of the time series matrix generated from EVG.
	
	\begin{myDef}
		The combination of EVG module and ZIP module constitutes a Deep Visibility Series (DVS) module.
	\end{myDef}
	
	\begin{figure}[htbp]
		\centerline{\includegraphics[scale=0.5]{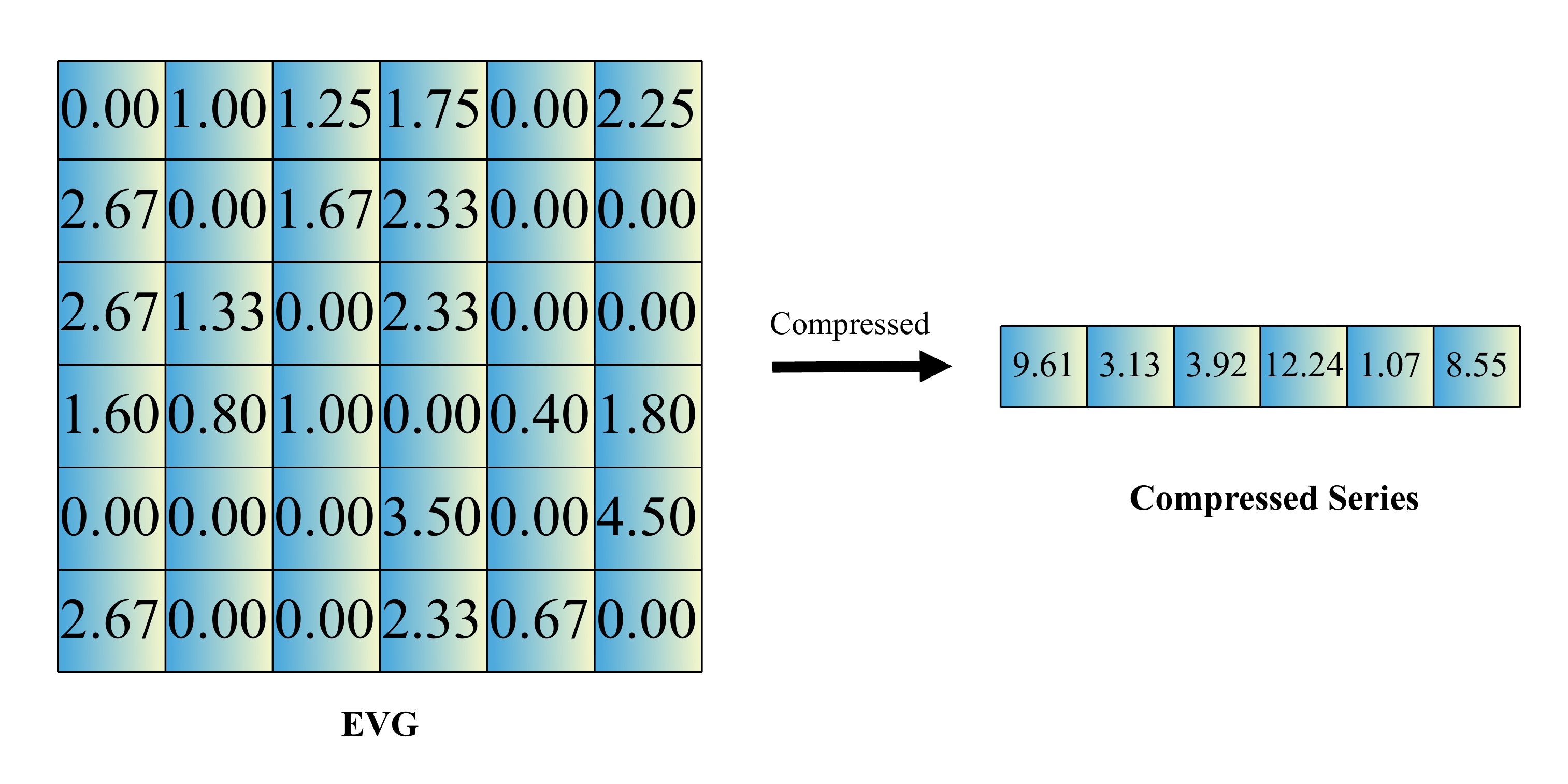}}
		\caption{Example of compressing process ($T = \left\{(1, 8),(2, 4),(3, 5),(4, 7),(5, 2),(6, 9)\right\}$)}
	\end{figure}
	
	\subsubsection{How to forecast}
	
	In the compression process, although the boundary information between time nodes is lost, the visible features are retained. Here, CNN is used to receive the sequence generated by DVS for deep learning. Compared with ANN, CNN can effectively extract the required features in complex information, which is represented here as the required time series information and visual network information. ANN, CNN, Long Short-Term Memory (LSTM) Network can all be used as training networks for DVS. The algorithm of DVS is as follows.
	
	\begin{algorithm}[htbp]
		\caption{$Deep\ Visibility\ Series\ Forecasting$}
		\begin{algorithmic}[1]
			\item Calculate matrix $A$ as:
			\For{Each time point $T_i$}
			\For{Each time point $T_j$}
			\item  $$A_{ij} = visual(T_i,T_j)$$
			\EndFor
			\EndFor
			\item Calculate matrix $B$ as:
			\For{Each time point $T_i$}
			\For{Each time point $T_j$}
			\item $$B_{ij} = \frac{A_{ij}*v_j}{Degree(T_i)}$$
			\EndFor
			\EndFor
			\item Calculate vector $zip$ as:
			\For{Each time point $T_i$}
			\item $$zip_i = \sum_{i=1}^{n} B_{ij}$$
			\EndFor
			\State  $$zip \xrightarrow{input} CNN \xrightarrow[test]{train} \hat{y}_{n+1}	$$			
			\Return  $\hat{y}_{n+1}$	
		\end{algorithmic}
	\end{algorithm}

	\section{Application in Construction Cost Index Forecasting}
	
	\subsection{Data set information}
	This experiment uses the CCI data set. The CCI data set has 295 construction cost data values, and the time includes the period from January 1990 to July 2014. The experimental CCI data set is processed, and the time series are split with a window size of 30 to generate 264 sets of sequences \cite{shahandashti2013forecasting,hwang2011time}. The ratio of training set to test set in the experiment is 8:2.
	
	\subsection{Model parameters}
	
	In the experimental part, the network structure used this time is shown in Fig.7.
	\begin{figure}[htbp]
		\centerline{\includegraphics[scale=0.6]{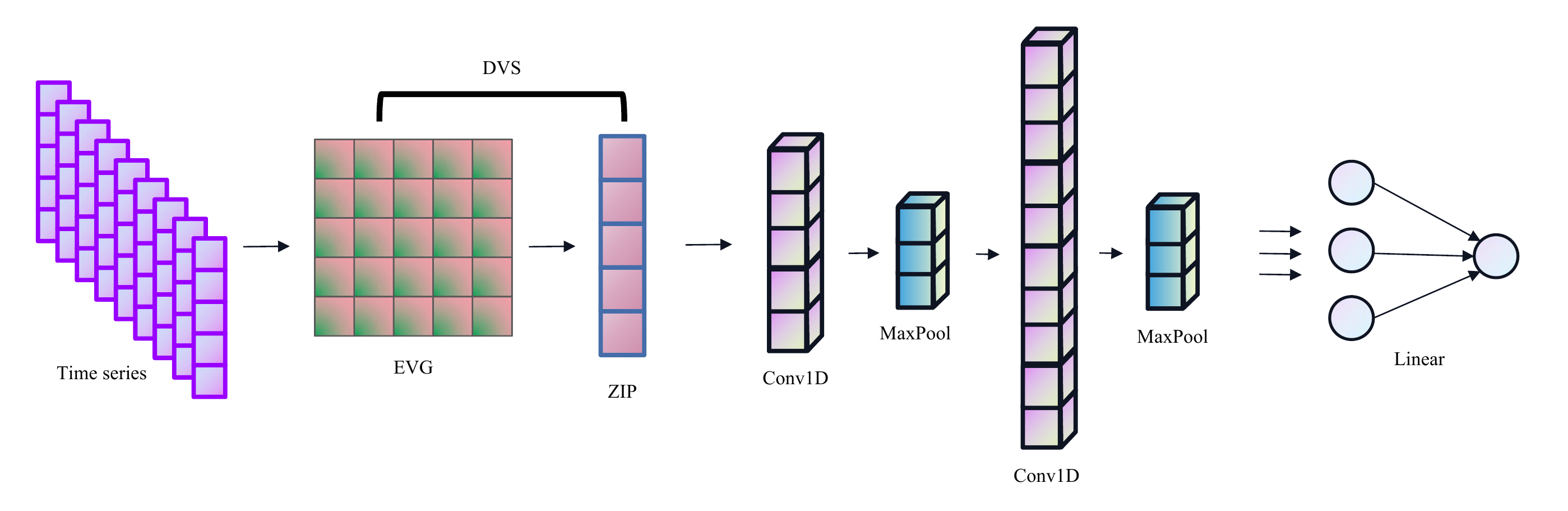}}
		\caption{Example of VG ($T = \left\{(1, 8),(2, 4),(3, 5),(4, 7),(5, 2),(6, 9)\right\}$)}
	\end{figure}
	
	The network used in the experiment is composed of DVS module and CNN, where CNN has two convolutional layers. The network used in the experiment uses the Cyclic Learning Rate (CLR) algorithm to dynamically adjust the learning rate while using the Adam optimizer to accelerate learning. Tab.1 shows the model parameters of the DVS+CNN network.
	
	\begin{table}[htbp]
		\setlength{\tabcolsep}{3mm}
		\centering
		\caption{DVS+CNN training network parameters}
		\begin{tabular}{cccccc}
			\hline
			& Structure            & \begin{tabular}[c]{@{}c@{}}Training\\ Iteration\end{tabular} & \begin{tabular}[c]{@{}c@{}}Learning\\ Rate\end{tabular} & \begin{tabular}[c]{@{}c@{}}Loss\\ Function\end{tabular} & \begin{tabular}[c]{@{}c@{}}Optimization\\ Function\end{tabular} \\ \hline
			& DVS                  &                                                              &                                                         &                                                         &                                                                 \\
			& Convolution Layer 1: &                                                              &                                                         &                                                         &                                                                 \\
			& In-Channel = 1       &                                                              &                                                         &                                                         &                                                                 \\
			& Kernel Size = 3      &                                                              &                                                         &                                                         &                                                                 \\
			& Out-Channel = 8      &                                                              &                                                         &                                                         &                                                                 \\
			& Pooling Layer1:      &                                                              &                                                         &                                                         &                                                                 \\
			& Pool Size=2          &                                                              &                                                         &                                                         &                                                                 \\
			& Convolution Layer 2: &                                                              &                                                         &                                                         &                                                                 \\
			DVS+CNN & In-Channel = 8       & 100                                                       & \begin{tabular}[c]{@{}c@{}}CLR\\ ($10^{-4},10^{-12}$)\end{tabular}                                            & MSE                                                     & Adam                                                            \\
			& Kernel Size = 3      &                                                              &                                                         &                                                         &                                                                 \\
			& Out-Channel = 16     &                                                              &                                                         &                                                         &                                                                 \\
			& Pooling Layer 2:     &                                                              &                                                         &                                                         &                                                                 \\
			& Pool Size=2          &                                                              &                                                         &                                                         &                                                                 \\
			& Flatten Layer        &                                                              &                                                         &                                                         &                                                                 \\
			& Linear Layer         &                                                              &                                                         &                                                         &                                                                 \\ \hline
		\end{tabular}
		
	\end{table}

	\subsection{Experimental results}
	
	\subsubsection{Experimental error standard}
	To evaluate the forecasting of DVS+CNN and comparison method, five measures of error is error standard in the experiment: mean absolute difference (MAD) \cite{yitzhaki2003gini} , mean absolute percentage error (MAPE) \cite{de2016mean} , symmetric mean absolute percentage error (SMAPE) \cite{tofallis2015better}, root mean square error (RMSE) \cite{hyndman2006another} , and normalized root mean squared error (NRMSE) \cite{shcherbakov2013survey} .
	
	\begin{equation}
		MAD=\frac{1}{N}\sum_{t=1}^{N}\left|\hat{y}(t)-y(t)\right| 
	\end{equation}
	\begin{equation}
		MAPE=\frac{1}{N}\sum_{t=1}^{N}\frac{\left|\hat{y}(t)-y(t)\right|}{y(t)}
	\end{equation}
	\begin{equation}
		SMAPE=\frac{2}{N}\sum_{t=1}^{N}\frac{\left|\hat{y}(t)-y(t)\right|}{\hat{y}(t)+y(t)}
	\end{equation}
	\begin{equation}
		RMSE=\sqrt{\frac{1}{N}\sum_{t=1}^{N}\left|\hat{y}(t)-y(t)\right|^{2}}
	\end{equation}
	\begin{equation}
		NRMSE=\frac{\sqrt{\frac{1}{N}\sum_{t=1}^{N}\left|\hat{y}(t)-y(t)\right|^{2}}}{y_{max}-y_{min}} 
	\end{equation}
	
	where $\hat y(t)$ is the predicted value, $y(t)$ is the true value and N is the total number of $\hat y(t)$.
	
	\subsubsection{Deep Visibility Series forecasting result}
	
	After dividing the CCI data set into a training set and a test set, the DVS+CNN network trains the CCI training set. After 100 iterations, the CCI test set is used as input for forecasting. Fig.8 is the forecasting effect of DVS+CNN on the test set. The forecasting result of DVS+CNN is close to the actual value of the CCI training set. At the same time, the forecasting value of DVS+CNN is stable and there is no jitter.
	
	\begin{figure}[htbp]
		\centerline{\includegraphics[scale=0.9]{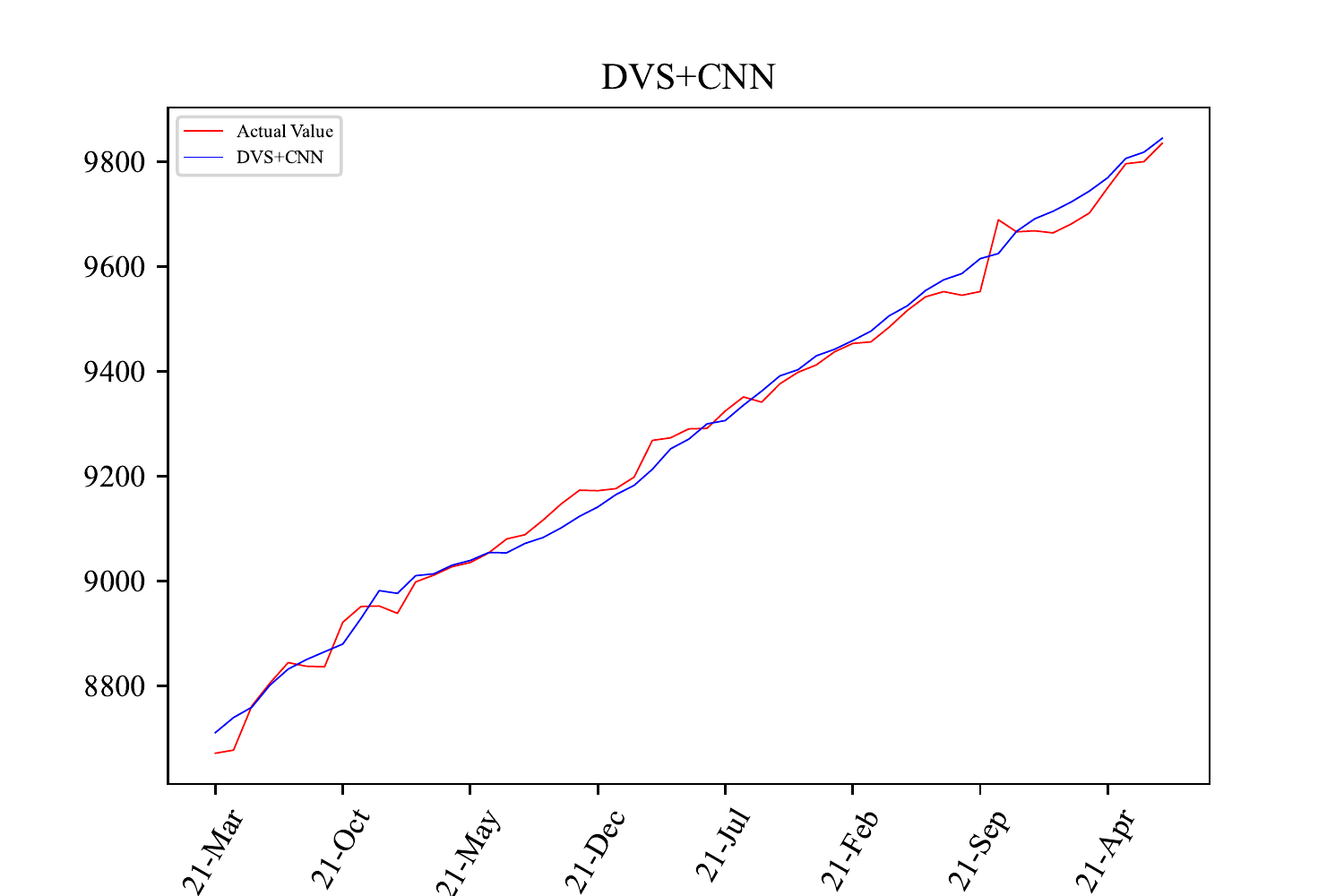}}
		\centerline{\includegraphics[scale=0.9]{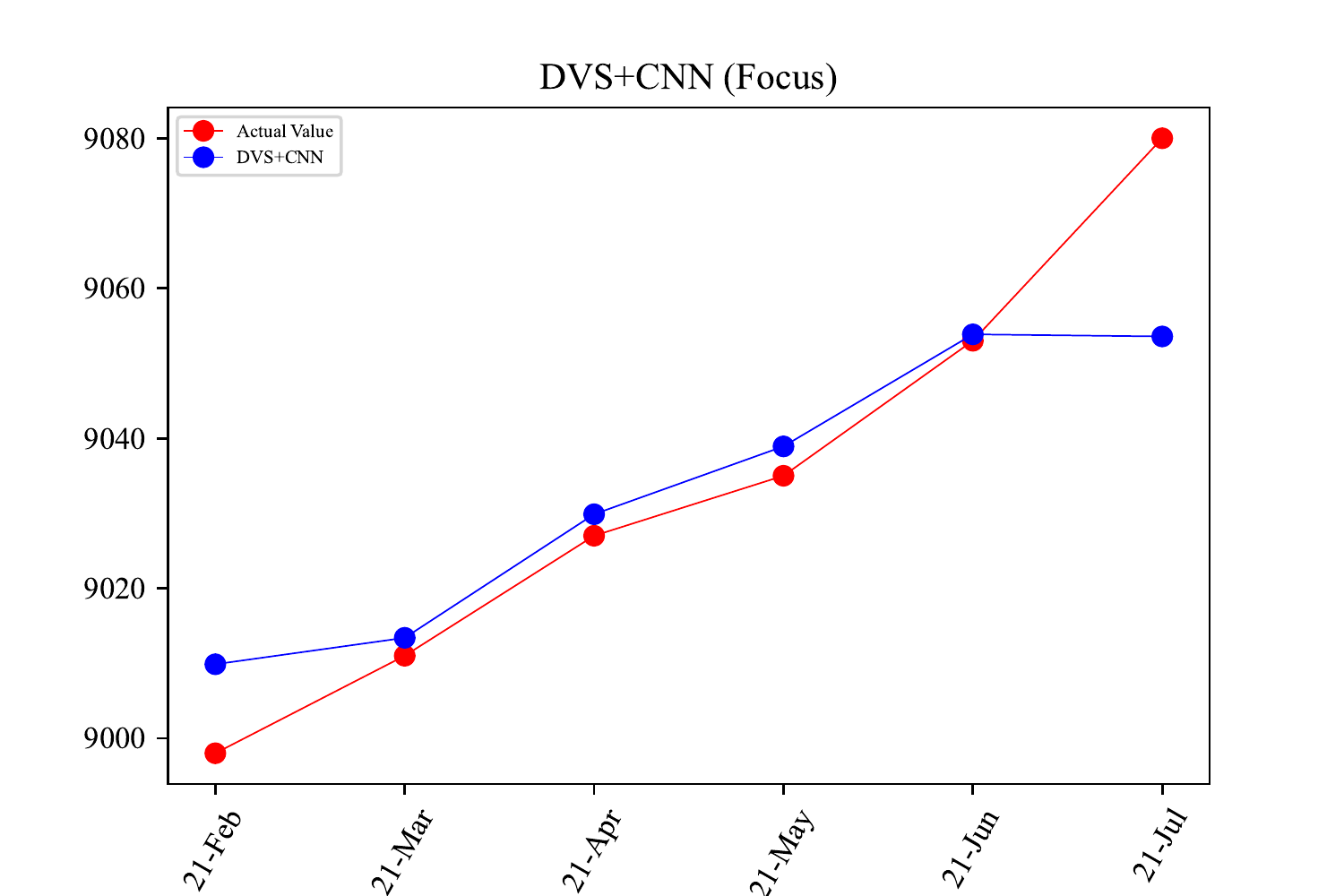}}
		\caption{DVS+CNN forecasting curve and CCI actual value}
	\end{figure}
	
	\subsubsection{Comparison of previous research and classical forecasting method}
	
	In terms of comparison methods, the experiment selected multiple methods from five different perspectives for comparison. The five perspectives are: previous perspectives, statistical methods, machine learning methods, hybrid methods and ablation experiments.
	
	Mao and Xiao's Method is a prediction method using VG in previous research \cite{mao2019time}. According to the introduction of past research, Mao and Xiao's method adopts a method of weighting the results of a pair of points linear fitting according to the distance.
	
	\begin{figure}[htbp]
		\centerline{\includegraphics[scale=0.9]{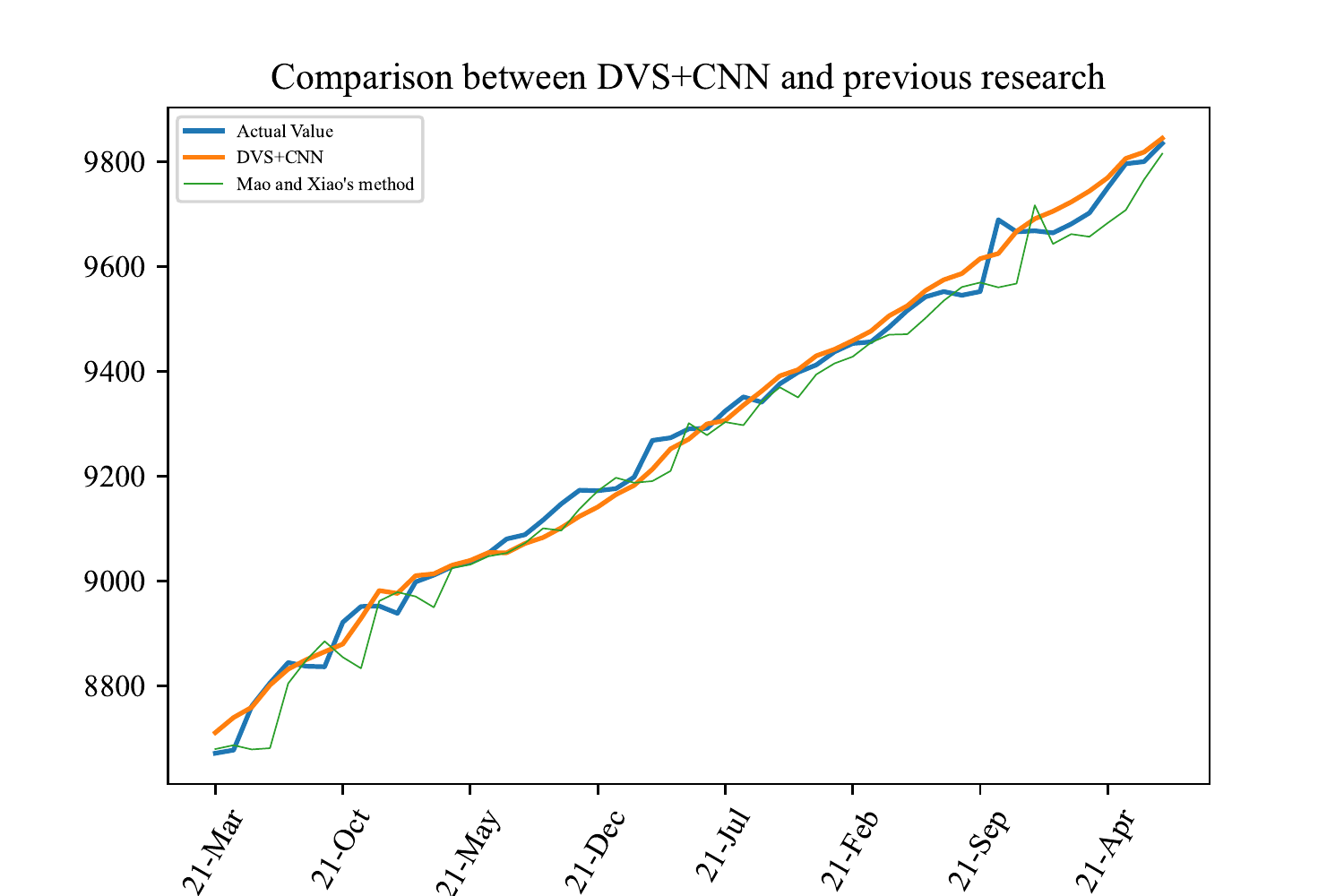}}
		\centerline{\includegraphics[scale=0.9]{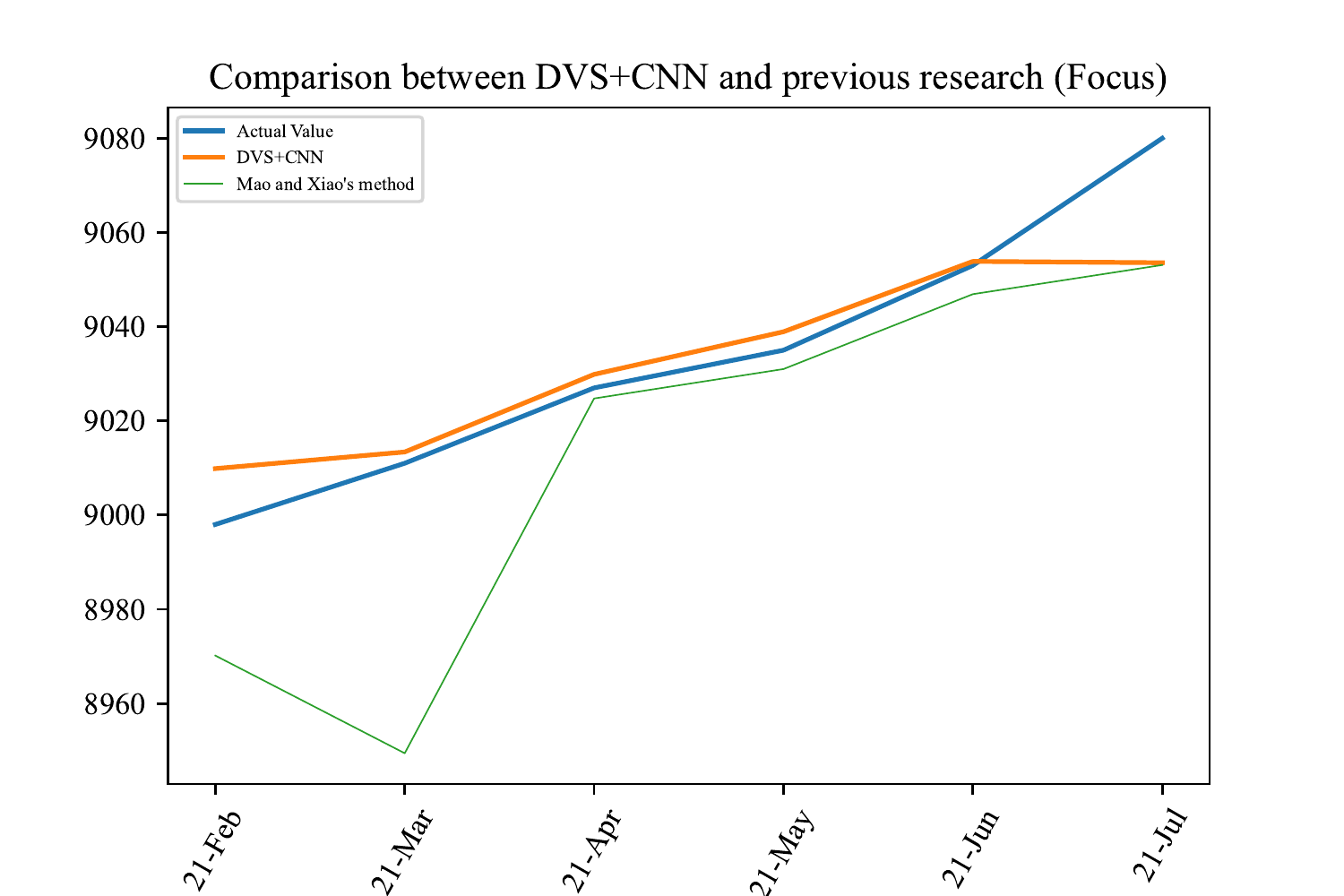}}
		\caption{DVS+CNN forecasting curve and previous research forecasting curve}
	\end{figure}
	
	Fig.9 shows the predicted value of DVS+CNN, the predicted value of Mao and Xiao's method and the actual value of CCI. Different from the prediction method in the previous study, the prediction of DVS+CNN is closer to the actual value, and the prediction result of DVS+CNN is less jitter than the previous study. The specific forecasting error between DVS+CNN and previous research methods is shown in Tab.2.
	
	\begin{table}[htbp]
		\setlength{\tabcolsep}{5mm}
		\centering
		\caption{Forecasting error between DVS+CNN and previous research methods}
		\begin{tabular}{cccccc}
			\hline
			& MAD              & MAPE            & SMAPE           & RMSE             & NRMSE            \\ \hline
			\textbf{DVS+CNN}      & \textbf{22.8781} & \textbf{0.2467} & \textbf{0.2468} & \textbf{28.4596} & \textbf{84.5042} \\
			Mao and Xiao's method & 36.2932          & 0.3936          & 0.3924          & 48.5065          & 143.8612         \\ \hline
		\end{tabular}
		
	\end{table}
	
	For the comparison between DVS+CNN and statistical methods, the statistical methods are selected: Simple Moving Average (SMA) (K=1) \cite{guan2017two}, Autoregressive Integrated Moving Average model (ARIMA) \cite{tseng2002combining}, Seasonal Autoregressive Integrated Moving Average model (Seasonal ARIMA) \cite{tseng2002fuzzy} and ExponenTial Smoothing (ETS) \cite{billah2006exponential} .
	
	\begin{figure}[htbp]
		\centerline{\includegraphics[scale=0.9]{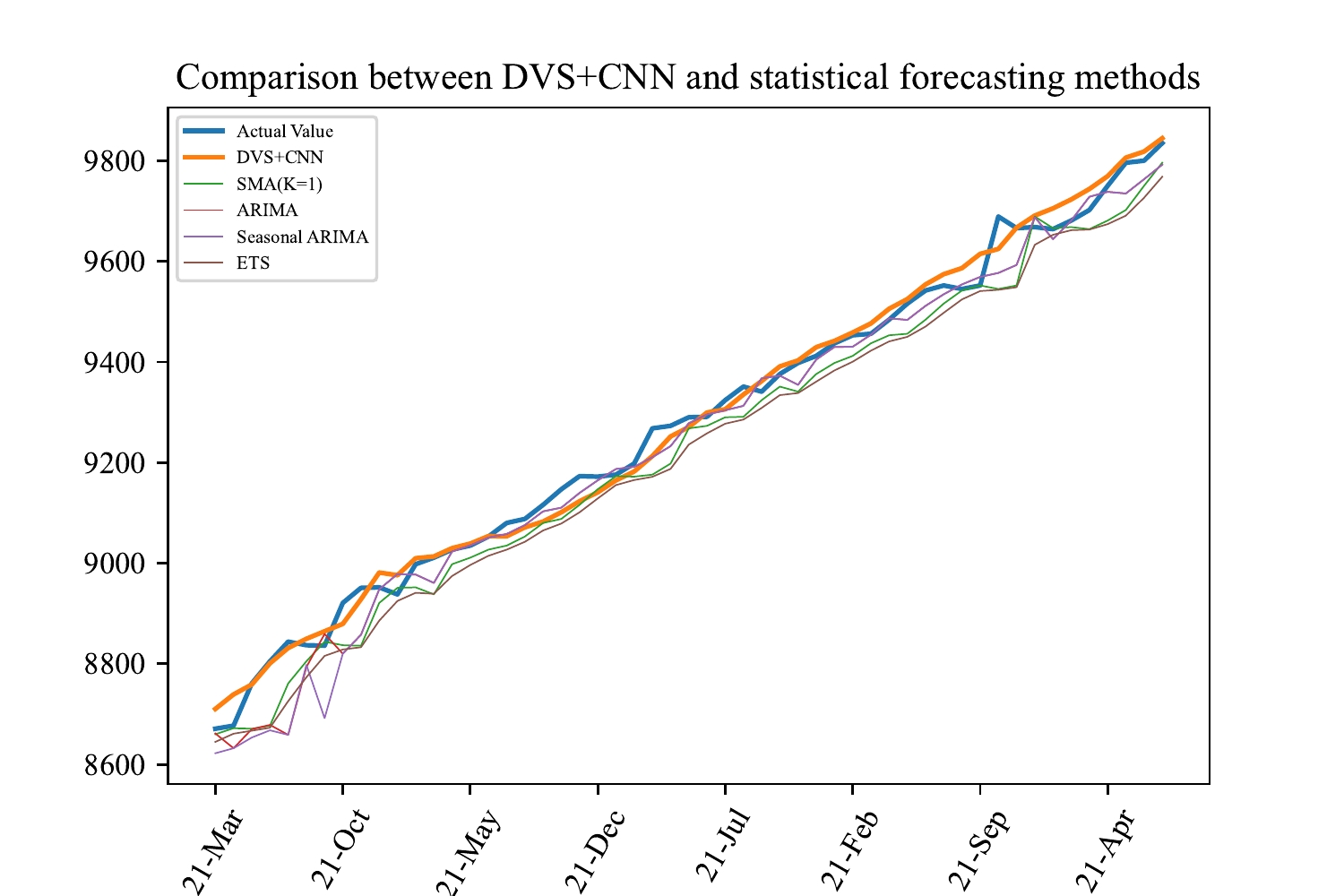}}
		\centerline{\includegraphics[scale=0.9]{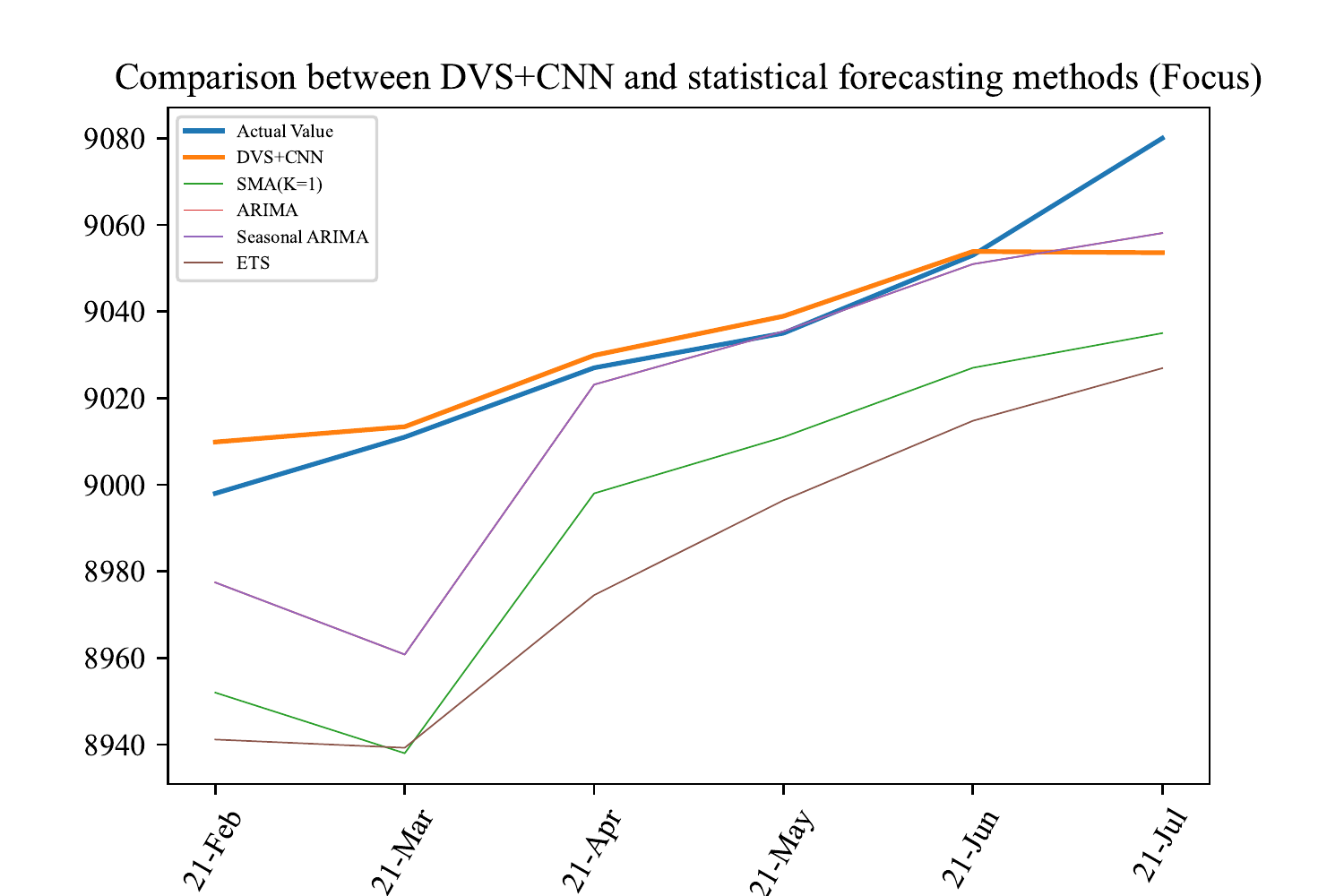}}
		\caption{DVS+CNN forecasting curve and statistical method forecasting curve}
	\end{figure}
	
	Fig.10 shows the predicted value of DVS+CNN, the predicted value of statistical methods and the actual value of CCI. The statistical method and DVS+CNN have less jitter, and the trend of the predicted curve is close to the actual curve. But compared with statistical methods, DVS+CNN is closer. The specific forecasting error between DVS+CNN and statistical methods is shown in Tab.3.
	
	\begin{table}[htbp]
		\centering
		\setlength{\tabcolsep}{5mm}
		\caption{Forecasting error between DVS+CNN and statistical methods}
		\begin{tabular}{cccccc}
			\hline
			& MAD              & MAPE            & SMAPE           & RMSE             & NRMSE            \\ \hline
			\textbf{DVS+CNN} & \textbf{22.8781} & \textbf{0.2467} & \textbf{0.2468} & \textbf{28.4596} & \textbf{84.5042} \\
			ARIMA            & 33.5646          & 0.3683          & 0.3669          & 49.401           & 145.05           \\
			Seasonal ARIMA   & 37.0188          & 0.4084          & 0.4066          & 54.5338          & 159.4385         \\
			SMA(K=1)         & 45.1132          & 0.4905          & 0.4886          & 56.2097          & 166.7715         \\
			ETS              & 57.8689          & 0.6299          & 0.6273          & 65.8998          & 196.6236         \\ \hline
		\end{tabular}
		
	\end{table}

	For the comparison between DVS+CNN and machine learning methods, the machine learning methods are selected:  Decision Tree Regression (DTR) \cite{hastie2009elements}, Ordinary least squares Linear Regression (Linear) \cite{hutcheson2011ordinary}, Lasso model fit with Least Angle Regression (Lasso) \cite{taylor2014post}, Support Vector Machines Regression (SVM) \cite{tong2009analysis}, Bayesian Ridge Regression (Bayesian) \cite{xu2020blood} and Logistic Regression (Logistic) \cite{hosmer2013applied,defazio2014saga} .
	
	\begin{figure}[htbp]
		\centerline{\includegraphics[scale=0.9]{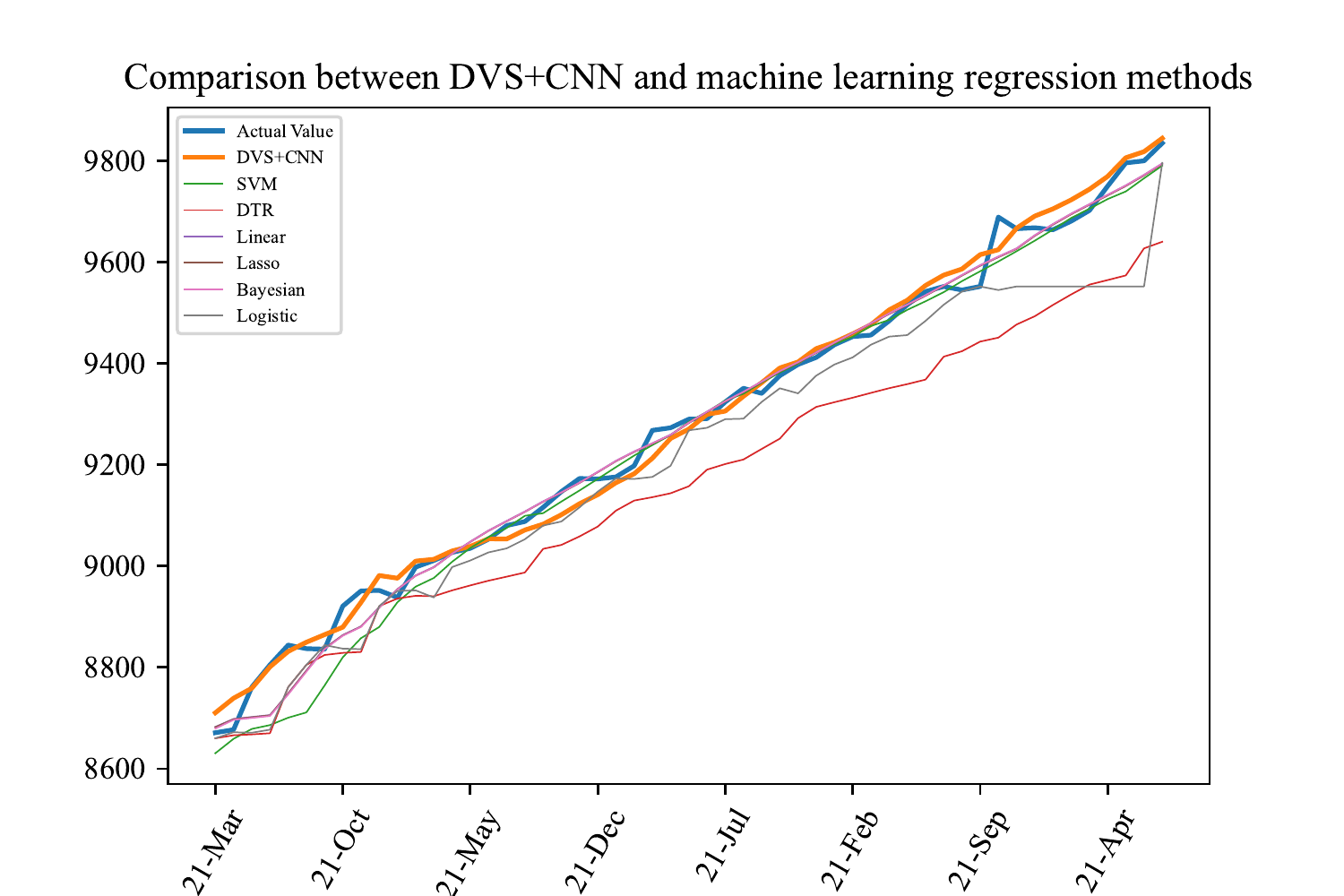}}
		\centerline{\includegraphics[scale=0.9]{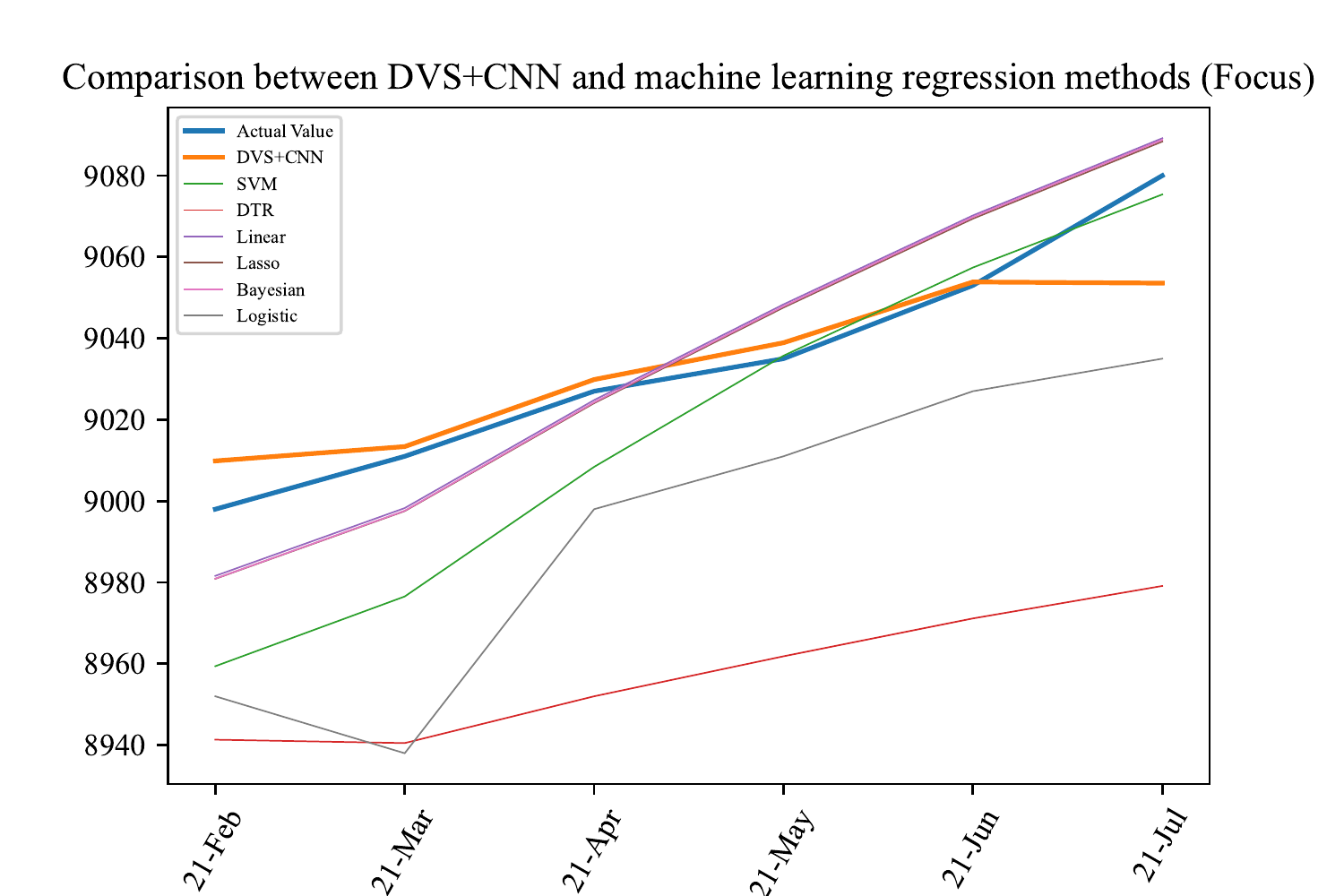}}
		\caption{DVS+CNN forecasting curve and machine learning method forecasting curve}
	\end{figure}
	
	Fig.11 shows the predicted value of DVS+CNN, the predicted value of machine learning methods and the actual value of CCI. Machine learning methods are not as stable as statistical methods. For example, the prediction effect of DTR has a large error. Some methods of machine learning have low prediction errors, such as Lasso. Compared with machine learning methods, DVS+CNN has the highest prediction accuracy and strong stability. The specific forecasting error between DVS+CNN and machine learning method is shown in Tab.4.
	
	\begin{table}[htbp]
		\centering
		\setlength{\tabcolsep}{5mm}
		\caption{Forecasting error between DVS+CNN and machine learning methods}
		\begin{tabular}{cccccc}
			\hline
			& MAD              & MAPE            & SMAPE           & RMSE             & NRMSE            \\ \hline
			\textbf{DVS+CNN} & \textbf{22.8781} & \textbf{0.2467} & \textbf{0.2468} & \textbf{28.4596} & \textbf{84.5042} \\
			Lasso            & 23.2985          & 0.2542          & 0.2537          & 32.7596          & 98.2587          \\
			Linear           & 23.3821          & 0.255           & 0.2546          & 32.5878          & 97.7052          \\
			Bayesian         & 23.4487          & 0.2558          & 0.2553          & 32.9991          & 98.7953          \\
			SVM              & 30.8839          & 0.3412          & 0.3399          & 46.6137          & 136.8565         \\
			Logistic         & 62.283           & 0.6711          & 0.6671          & 84.0805          & 249.4629         \\
			DTR              & 111.3944         & 1.2038          & 1.1952          & 122.5928         & 391.5623         \\ \hline
		\end{tabular}
	\end{table}

	For the comparison between DVS+CNN and hybrid methods, the statistical method is selected: ARIMA+ANN \cite{babu2014moving} and ETS+ANN \cite{panigrahi2017hybrid} . Hybrid forecasting methods usually combine a statistical forecasting method with a deep learning method. In the hybrid method, the deep learning network generally chooses ANN to process the residuals in the statistical prediction, and then completes the prediction through synthesis, which has a lower prediction error. DVS+CNN is also an optimized time series prediction method. DVS optimizes the VG algorithm. By comparing DVS+CNN and hybrid methods, it can effectively show the advantages of DVS+CNN.
	
	\begin{figure}[htbp]
		\centerline{\includegraphics[scale=0.9]{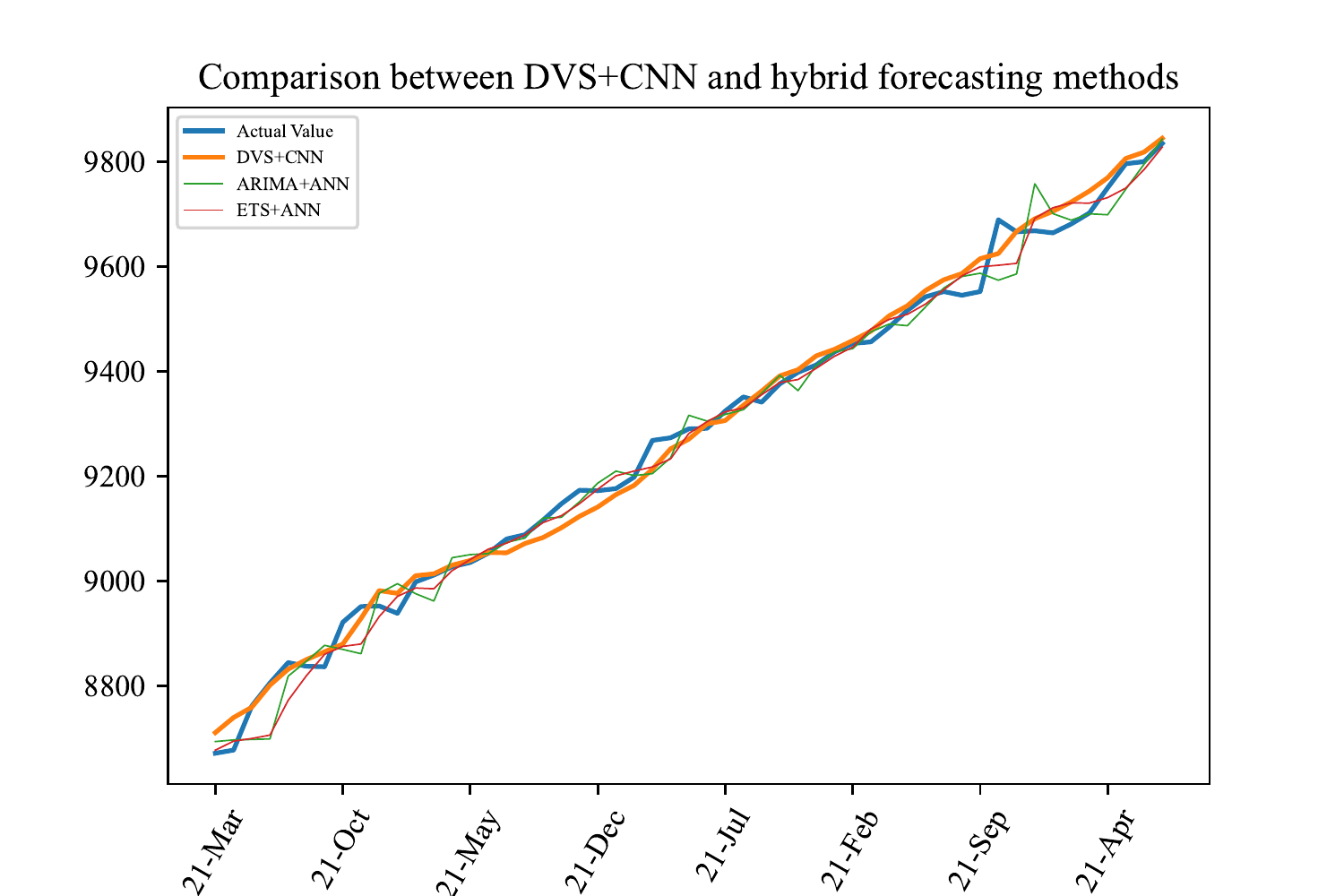}}
		\centerline{\includegraphics[scale=0.9]{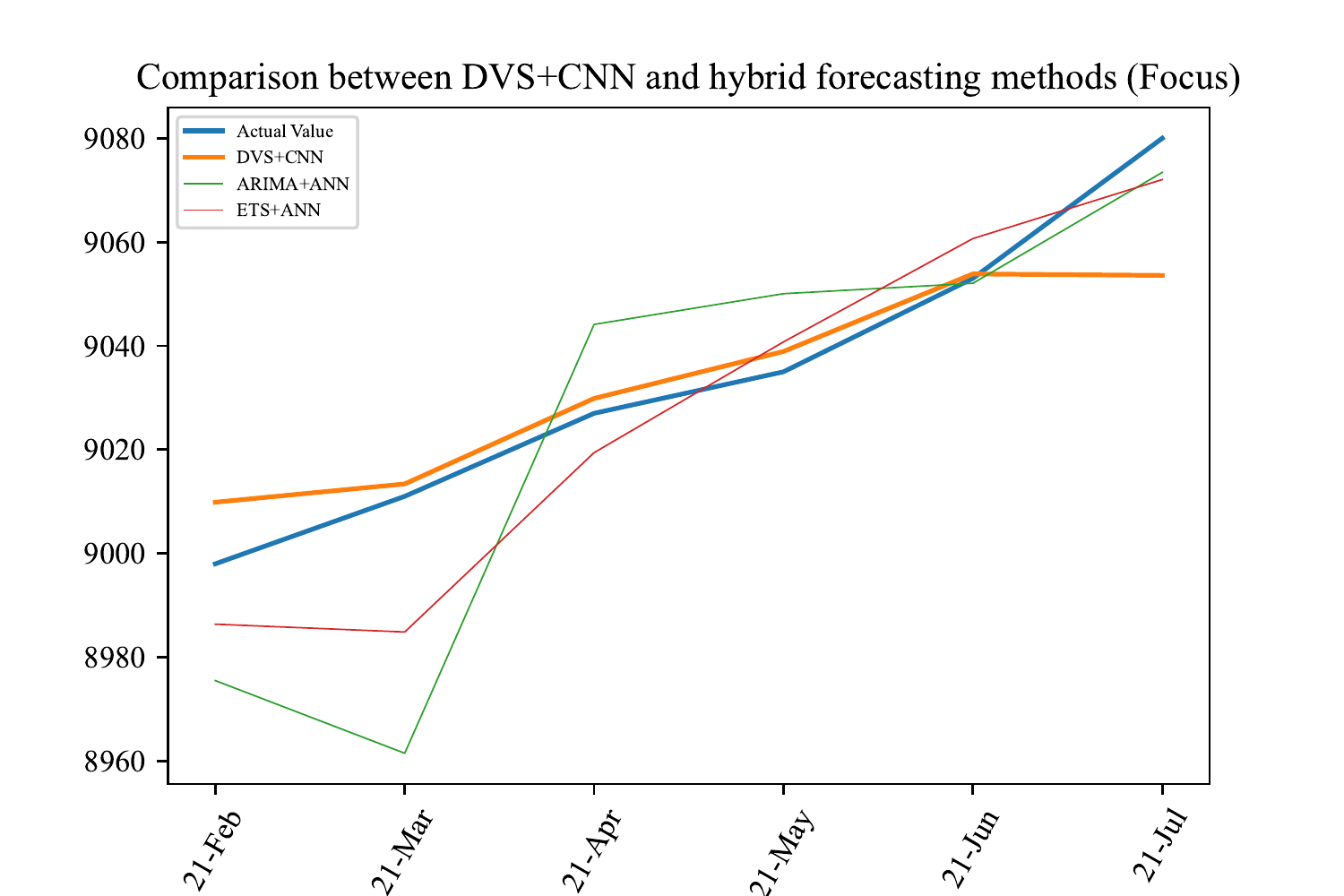}}
		\caption{DVS+CNN forecasting curve and comparison method forecasting curve}
	\end{figure}
	
	Fig.12 shows the predicted value of DVS+CNN, the predicted value of Mao and Xiao's method and the actual value of CCI. Compared with pure ARIMA and ETS, the hybrid method further reduces the prediction error. Although the hybrid method has a lower prediction error, compared with DVS+CNN, DVS+CNN has higher prediction accuracy. The specific forecasting error between DVS+CNN and hybrid  methods is shown in Tab.5.
	
	\begin{table}[htbp]
		\setlength{\tabcolsep}{5mm}
		\centering
		\caption{Forecasting error between DVS+CNN and hybrid methods}
		\begin{tabular}{cccccc}
			\hline
			& MAD              & MAPE            & SMAPE           & RMSE             & NRMSE            \\ \hline
			\textbf{DVS+CNN} & \textbf{22.8781} & \textbf{0.2467} & \textbf{0.2468} & \textbf{28.4596} & \textbf{84.5042} \\
			ETS+ANN          & 25.0547          & 0.2719          & 0.2715          & 33.8315          & 99.7235          \\
			ARIMA+ANN        & 29.7483          & 0.3226          & 0.3221          & 40.4425          & 119.2806         \\ \hline
		\end{tabular}
		
	\end{table}

	By comparing previous studies, statistical methods, machine learning methods and statistical methods, the superiority of DVS+CNN has been demonstrated. In order to rigorously further illustrate the superiority of DVS, the experimental part includes ablation experiment. In the ablation experiment, DVS and LSTM, CNN and ANN are separated, and the effect of the DVS module is proved by comparing the experimental errors between the DVS module and the non-DVS module. In the experiment, the model parameters of ANN \cite{borghi2021covid} , CNN \cite{zeng2016self} and LSTM \cite{cao2019financial} are shown in Tab.6.
	
	\begin{table}[htbp]
		\centering
		\setlength{\tabcolsep}{5mm}
		\caption{Comparison of method parameters for ablation experiments}
		\begin{tabular}{cccccc}
			\hline
			& \begin{tabular}[c]{@{}c@{}}Structure\end{tabular} & \begin{tabular}[c]{@{}c@{}}Training \\ Iteration\end{tabular} & \begin{tabular}[c]{@{}c@{}}Learning \\ Rate\end{tabular} & \begin{tabular}[c]{@{}c@{}}Loss\\ Function\end{tabular} & \begin{tabular}[c]{@{}c@{}}Optimizer\\ Function\end{tabular} \\ \hline
			& Hidden Layer=100                                                                                                                                    &                                                               &                                                          &                                                         &                                                              \\
			ANN      & Activation=Relu                                                                                                                                     & 100                                                         & 0.01                                                     & MSE                                                     & Adam                                                         \\
			& Output Layer=1                                                                                                                                      &                                                               &                                                          &                                                         &                                                              \\ \hline
			& Hidden Layer=50                                                                                                                                     &                                                               &                                                          &                                                         &                                                              \\
			& Activation=Relu                                                                                                                                     &                                                               &                                                          &                                                         &                                                              \\
			LSTM     & Input Timestep=3                                                                                                                                    & 100                                                         & 0.01                                                     & MSE                                                     & Adam                                                         \\
			& Output Timestep=1                                                                                                                                   &                                                               &                                                          &                                                         &                                                              \\
			& Dense Layer=1                                                                                                                                       &                                                               &                                                          &                                                         &                                                              \\ \hline
			& Convolution Layer:                                                                                                                                  &                                                               &                                                          &                                                         &                                                              \\
			& Filters=64                                                                                                                                          &                                                               &                                                          &                                                         &                                                              \\
			& Kernel Size=2                                                                                                                                       &                                                               &                                                          &                                                         &                                                              \\
			CNN      & Activation=Relu                                                                                                                                     & 100                                                         & 0.01                                                     & MSE                                                     & Adam                                                         \\
			& Pooling Layer:                                                                                                                                      &                                                               &                                                          &                                                         &                                                              \\
			& Pool Size=2                                                                                                                                         &                                                               &                                                          &                                                         &                                                              \\
			& Dense Layer=100                                                                                                                                     &                                                               &                                                          &                                                         &                                                              \\
			& Dense Layer=1                                                                                                                                       &                                                               &                                                          &                                                         &                                                              \\ \hline
		\end{tabular}
		
	\end{table}

	Fig.13 shows the predicted value of ablation experiment value and the actual value of CCI. Among them, because the LSTM network has a large prediction error when it is trained for 100 epochs, the images of LSTM and DVS+LSTM are not drawn in Fig.13. The specific forecasting error in the  ablation experiment is shown in Tab.5. The error of LSTM is much higher than other neural network prediction errors because 100 epoch is not enough for the training volume of LSTM network, and the training efficiency of LSTM is lower than that of ANN and CNN. Comparing the ANN and CNN using the DVS module with the ANN and CNN without the DVS module, the prediction error of the neural network using the DVS module is further reduced. Among them, DVS has the greatest improvement in the prediction accuracy of CNN networks. DVS+CNN has the best effect among all the comparison methods.
	
	\begin{figure}[htbp]
		\centerline{\includegraphics[scale=0.9]{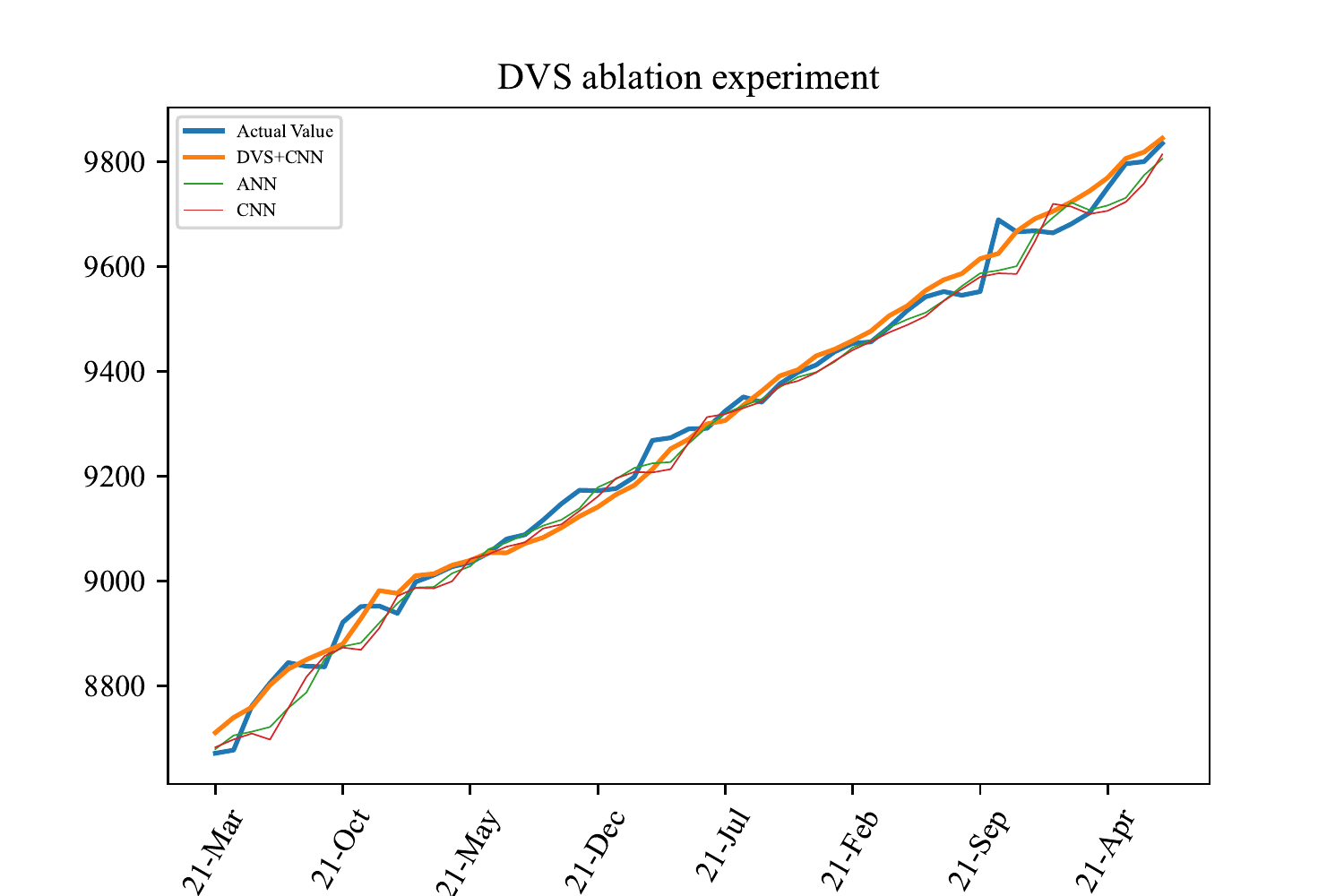}}
		\centerline{\includegraphics[scale=0.9]{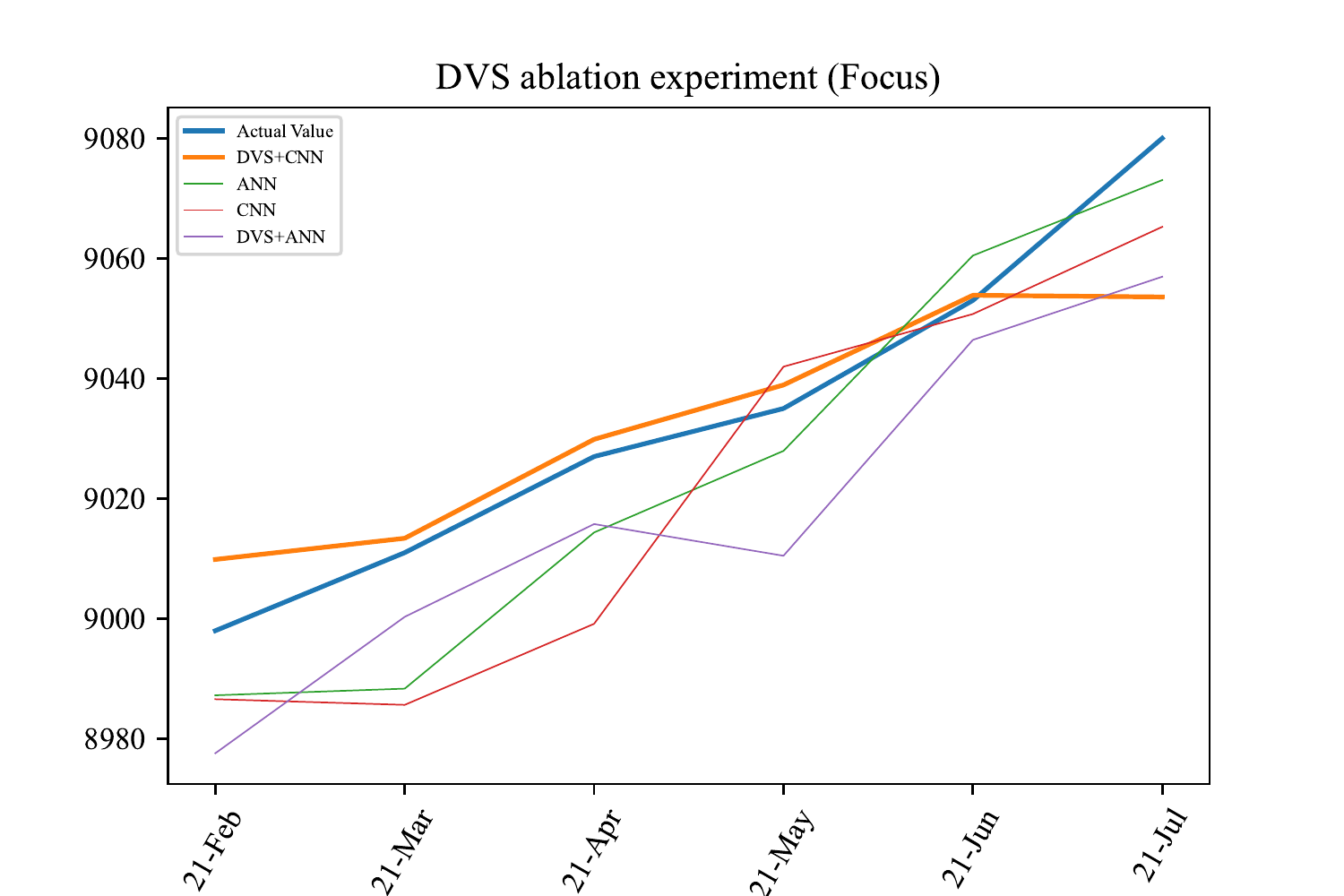}}
		\caption{DVS+CNN forecasting curve and comparison method forecasting curve}
	\end{figure}
	
	\begin{table}[htbp]
		\setlength{\tabcolsep}{5mm}
		\centering
		\caption{Forecasting error in the ablation experiments (DVS+LSTM has a large prediction error in small training epochs, which is indicated by "—")}
		\begin{tabular}{cccccc}
			\hline
			& MAD              & MAPE            & SMAPE           & RMSE             & NRMSE            \\ \hline
			\textbf{DVS+CNN} & \textbf{22.8781} & \textbf{0.2467} & \textbf{0.2468} & \textbf{28.4596} & \textbf{84.5042} \\
			DVS+ANN          & 24.6384          & 0.2674          & 0.2672          & 29.821           & 88.2419          \\
			ANN              & 26.3411          & 0.286           & 0.2854          & 35.0024          & 104.294          \\
			CNN              & 30.7757          & 0.334           & 0.3332          & 40.1989          & 119.5102         \\
			LSTM             & 376.1784         & 41944.6893      & 7.8407          & 1796.6998        & 1814.6961        \\
			DVS+LSTM         & -                & -               & -               & -                & -                \\ \hline
		\end{tabular}
		
	\end{table}

	\subsection{Analysis}
	
	\subsubsection{Comparative experiment analysis}
	This experiment contains a total of six sub-experiments. The comparison between the design of DVS+CNN and the previous research shows that DVS+CNN has greatly improved the previous research. The comparison of design statistical methods and machine learning methods with DVS+CNN shows that DVS+CNN has a good forecasting effect, which is better than traditional methods. The hybrid method is an improved method based on statistical methods. By comparing different improvement methods, the rationality of the improvement of DVS+CNN is explained. Finally, an ablation experiment is set up, which proves that the progress of DVS+CNN is due to the DVS module, which once again proves the superiority of the DVS module.
	
	\subsubsection{Deep Visibility Series superiority analysis}
	The third section in the article mentioned how DVS is designed. Like most neural network models, the design of DVS is also a bionic design. Although the original VG and random walk algorithm utilize the visible feature of nodes, in the end only appropriate nodes are selected for weighting, without involving the specific nature of the network, and it uses less potential information. DVS strengthens the information carried by the visible view through EVG, and uses the principle of biological vision to save more network information in the time series. The time complexity of DVS is $O(n^2)$ and the time complexity is equal to that of a simple neural network. DVS cooperates with CNN to extract effective network information and time series information. The neural network forecasting method without DVS module does not cover the potential network information when forecasting the time series leading to larger forecasting error.

	\section{Conclusion}
	
	In order to improve the previous research and optimize the forecasting method based on VG, the DVS module is proposed, combining the VG algorithm with CNN. The EVG module in the DVS solves the problem of incomplete utilization of the original VG and time series forecasting information, greatly improving the forecasting efficiency and reducing the forecasting error. At the same time, DVS prediction has a good forecasting effect on the CCI data set, which has contributed to CCI prediction. In future work, DVS will continue to improve by combining bionic design and using biological vision principles.

	\section*{Authorship contributions}
	\noindent \textbf{Tianxiang Zhan}: Conceptualization, Methodology, Software,
	Writing – original draft. \\
	\textbf{Yuanpeng He}: Visualization, Writing – review \& editing. \\
	\textbf{Hanwen Li}: Visualization, Writing – review \& editing. \\
	\textbf{Fuyuan Xiao}: Writing – review \& editing, Project administration, Funding acquisition, Supervision.
	
	\section*{Declarations}
	\noindent \textbf{Conflict of interest} Authors Tianxiang Zhan, Yuanpeng He, Fuyuan Xiao declare that they have no conflict of interest.
	\\
	\textbf{Funding} This study was funded by National Natural Science Foundation of China (No. 62003280) and Chongqing Talents: Exceptional Young Talents Project (CQYC202105031).
	\\
	\textbf{Ethical approval} This article does not contain any studies with human participants or animals performed by any of the authors.
	\\
	\textbf{Informed consent} Informed consent was obtained from all individual
	participants included in the study.

	\bibliographystyle{elsarticle-num}
	\bibliography{References}

\end{document}